%% file: arxiv_arr.tex
\pdfoutput=1

\documentclass[11pt]{article}

\usepackage[final]{sty/acl}

\usepackage{times}
\usepackage{latexsym}

\usepackage[T1]{fontenc}

\usepackage[utf8]{inputenc}

\usepackage{microtype}

\usepackage{inconsolata}

\usepackage{graphicx}

\usepackage{booktabs}       
\usepackage{amsfonts}       
\usepackage{amssymb}
\usepackage{nicefrac}       
\usepackage{microtype}      
\usepackage{xcolor}         
\usepackage{amsmath}
\usepackage{amsfonts,amssymb}
\usepackage{multirow}
\usepackage{booktabs}
\usepackage{hyperref}
\usepackage{float}
\usepackage{enumitem}
\usepackage{algorithmicx}
\usepackage{algpseudocode}
\usepackage{tabularx}
\usepackage{subcaption}
\usepackage{marvosym}
\usepackage{adjustbox} 
\usepackage{wrapfig}
\usepackage{algorithm}
\usepackage{algpseudocode}
\usepackage{textcomp}
\usepackage{fontawesome}
\usepackage{tcolorbox}
\usepackage{colortbl}

\input{math_commands}

\definecolor{darkblue}{rgb}{0, 0, 0.5}
\hypersetup{citecolor=darkblue}
\definecolor{lightred}{RGB}{255,230,230}
\definecolor{lightgreen}{RGB}{230,255,230}
\definecolor{mediumgreen}{RGB}{200,255,200}
\definecolor{darkgreen}{RGB}{150,255,150}
\definecolor{deepdarkgreen}{RGB}{0,120,0}
\definecolor{lightred}{RGB}{255,230,230}
\definecolor{promptgreen}{RGB}{145, 204, 117}
\definecolor{tagthink}{RGB}{57, 106, 177}
\definecolor{tagcall}{RGB}{218, 124, 48}
\definecolor{tagresponse}{RGB}{93, 156, 89}
\definecolor{taganswer}{RGB}{173, 87, 156}
\definecolor{tableline}{RGB}{166, 194, 226}
\definecolor{tablehead}{RGB}{226, 236, 249}
\definecolor{tablesection}{RGB}{240, 246, 252}
\definecolor{ourssection}{RGB}{229, 242, 224}
\definecolor{oursbase}{RGB}{249, 252, 248}
\definecolor{ourssft}{RGB}{239, 248, 235}
\definecolor{oursrl}{RGB}{220, 239, 211}

\newcommand{\tagtoken}[2]{\begingroup\setlength{\fboxsep}{1pt}\colorbox{#1!12}{\textcolor{#1}{\scriptsize\ttfamily #2}}\endgroup}
\newcommand{\thinktag}{\tagtoken{tagthink}{<think>}}
\newcommand{\toolcalltag}{\tagtoken{tagcall}{<tool\_call>}}
\newcommand{\toolresponsetag}{\tagtoken{tagresponse}{<tool\_response>}}
\newcommand{\answertag}{\tagtoken{taganswer}{<answer>}}
\newcommand{\tagclose}[2]{\tagtoken{#1}{</#2>}}
\newcommand{\tagrow}[2]{#1 & #2 \\}
\newcommand{\modelname}{\textsc{UniBrowse}}

\newtcolorbox{trajectorybox}{
colback=black!2!white,
colframe=black!35,
arc=3pt,
boxrule=0.35pt,
left=4pt,
right=4pt,
top=4pt,
bottom=4pt,
}

\newtcolorbox{promptbox}[2][Prompt]{
colback=black!5!white,
arc=5pt, 
boxrule=0.5pt,
fonttitle=\bfseries,
left=0pt,
right=0pt,
top=0pt,
bottom=0pt,
title=#1, 
before upper={\small}, fontupper=\fontfamily{ptm}\selectfont,
colframe=#2,
}

\title{\modelname: A Data-to-Agent Framework for Multimodal BrowseComp}

\author{Xiyu Wei$^{1,2}$\thanks{Equal contribution.}~~, Qingwei Zong$^{1,3}$\footnotemark[1]~~,Zhuocheng Yu$^{1,3}$\footnotemark[1]~~, Sujian Li$^{1,3}$~\thanks{Corresponding authors.}~\\
$^1$  Key Laboratory of Computational Linguistics, MOE,  Peking University \\
$^2$  School of Software and Microelectronics, Peking University\\
$^3$ School of Computer Science, Peking University \\
\texttt{ \{wxylemon, shiinasama\}@stu.pku.edu.cn \quad  lisujian@pku.edu.cn }\\
}

\begin{document}

\maketitle

\begin{figure*}[ht]
    \vspace{-20mm}
    \centering
    \includegraphics[width=1.0\textwidth]{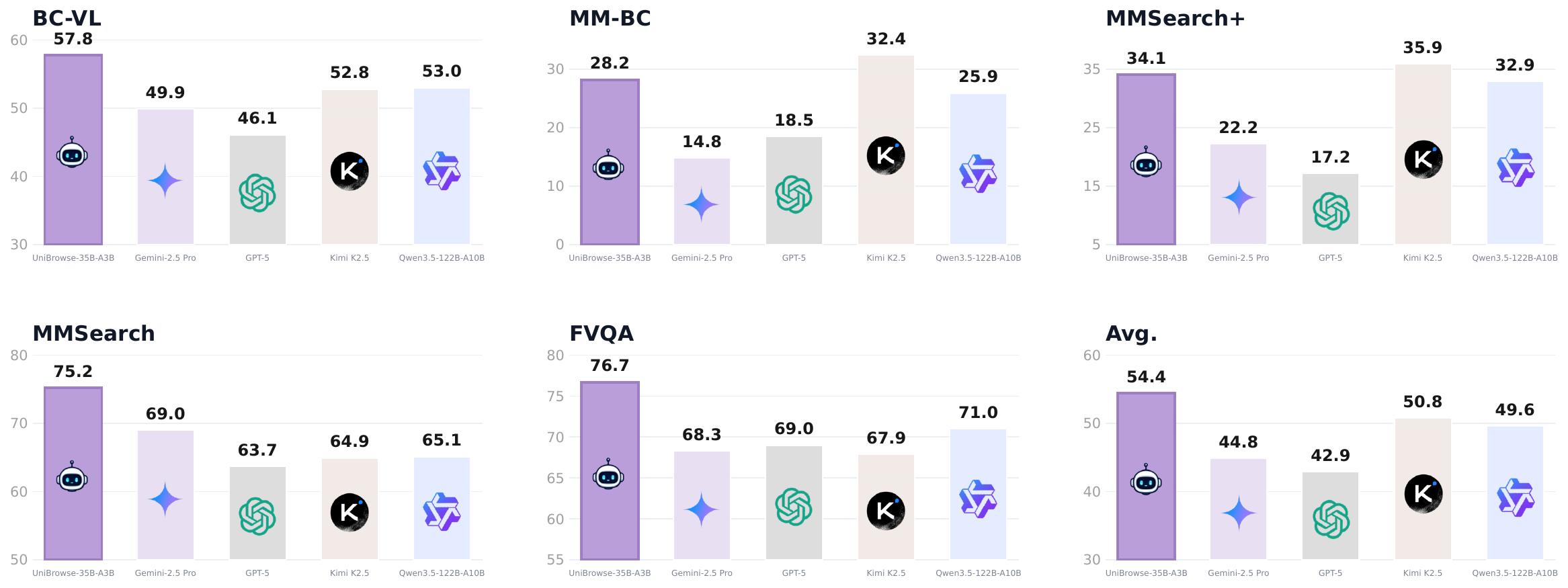}
    \caption{Overall performance of \modelname\ Agent compared to other models across five benchmarks.}
    \label{fig:main_results}
\end{figure*}

\begin{abstract}
Multimodal BrowseComp tasks require agents to combine perception, tool use, and long-horizon reasoning over dynamic web content, challenging their ability to handle compositional structure, open-world uncertainty, and multimodal integration across extended interactions.
Crucially, real-world multimodal browsing involves three distinct information-flow patterns: text-only, image-to-text, and text-to-image, yet existing data construction methods cover only the text-only and image-to-text patterns, leaving text-to-image largely unaddressed and limiting agent generality and robustness. We introduce \textbf{\modelname}, a unified data pipeline that for the first time simultaneously generates training data covering all three patterns, augments curated knowledge graphs with live web retrieval for improved fidelity, and introduces a novel metric of exploration degree to filter low-signal instances for efficient reinforcement learning.
Through this pipeline, we produce high-quality cold-start tool-use trajectories and exploration-rich QA pairs, and train a 35B-scale agent via supervised fine-tuning and exploration-aware RL. 
The resulting \modelname\ agent achieves state-of-the-art performance on multimodal BrowseComp benchmarks, attaining an average accuracy of 54.4 across five diverse benchmarks---an improvement of 10.5 points over its base model Qwen3.5-35B-A3B---and surpassing serveral closed-source agent workflows such as GPT-5 (42.9), Gemini-2.5 Pro (44.8), and Gemini-2.5 Flash (41.3).
\end{abstract}

\section{Introduction}


BrowseComp-style tasks evaluate whether web browsing agents can solve questions whose answers are difficult to locate and must be composed from multiple pieces of web evidence~\citep{wei2025browsecomp}. Unlike standard fact-seeking tasks~\citep{nakano2021webgpt}, these problems cannot be solved by issuing a single query or extracting a single webpage snippet. Instead, an agent must plan searches, follow partial clues, assess evidence reliability, and combine information across sources to arrive at a uniquely determined answer. Real-world browsing further extends this challenge beyond text: users often provide image clues, ask agents to identify visually grounded entities, or require visual verification after textual search has narrowed down the target. This gives rise to the more challenging setting of \emph{multimodal BrowseComp}, where agents must jointly perform textual retrieval, visual grounding, tool use, and long-horizon reasoning over open-web evidence. 
This emerging setting has therefore attracted increasing attention from the research community~\citep{jiang2024mmsearch,li2025mm}.


Despite this growing interest, constructing effective training data for multimodal BrowseComp remains difficult.
Early steps toward injecting visual information into the data generation process include WebWatcher~\citep{geng2025webwatcher}, which relies on curated knowledge graphs to obtain structured relations, as well as Vision‑DeepResearch~\citep{zeng2026vision} and Skywork R1V4~\citep{zhang2025skywork}, which employ web random walks to mimic authentic browsing environments. Although these pioneering methods succeed in producing large-scale multimodal training instances, their further application is hindered by a lack of task diversity and insufficient training efficiency, which manifests in three key aspects.
First, the existing dataset exclusively covers the \emph{image-to-text} pattern where an initial image clue is resolved into textual evidence, but misses the equally important \emph{text-to-image} pattern where textual browsing precedes visual retrieval. This leaves agents undertrained for scenarios requiring visual-after-textual reasoning. 
Second, 
existing methods rely exclusively on a single construction paradigm, forcing an all-or-nothing trade-off: graph-based methods suffer from restricted web diversity, while random-walk methods expose data generation to massive noise and false evidence. 
Third, 
prior works apply learning directly on the generated data without a filtering mechanism tailored for the long‑horizon, multi‑step nature of BrowseComp reasoning. Consequently, low‑quality trajectories are fed unfiltered into training, wasting computational resources and slowing down model training.

\begin{figure*}[t]
    \vspace{-20mm}
    \centering
    \includegraphics[width=1.0\textwidth]{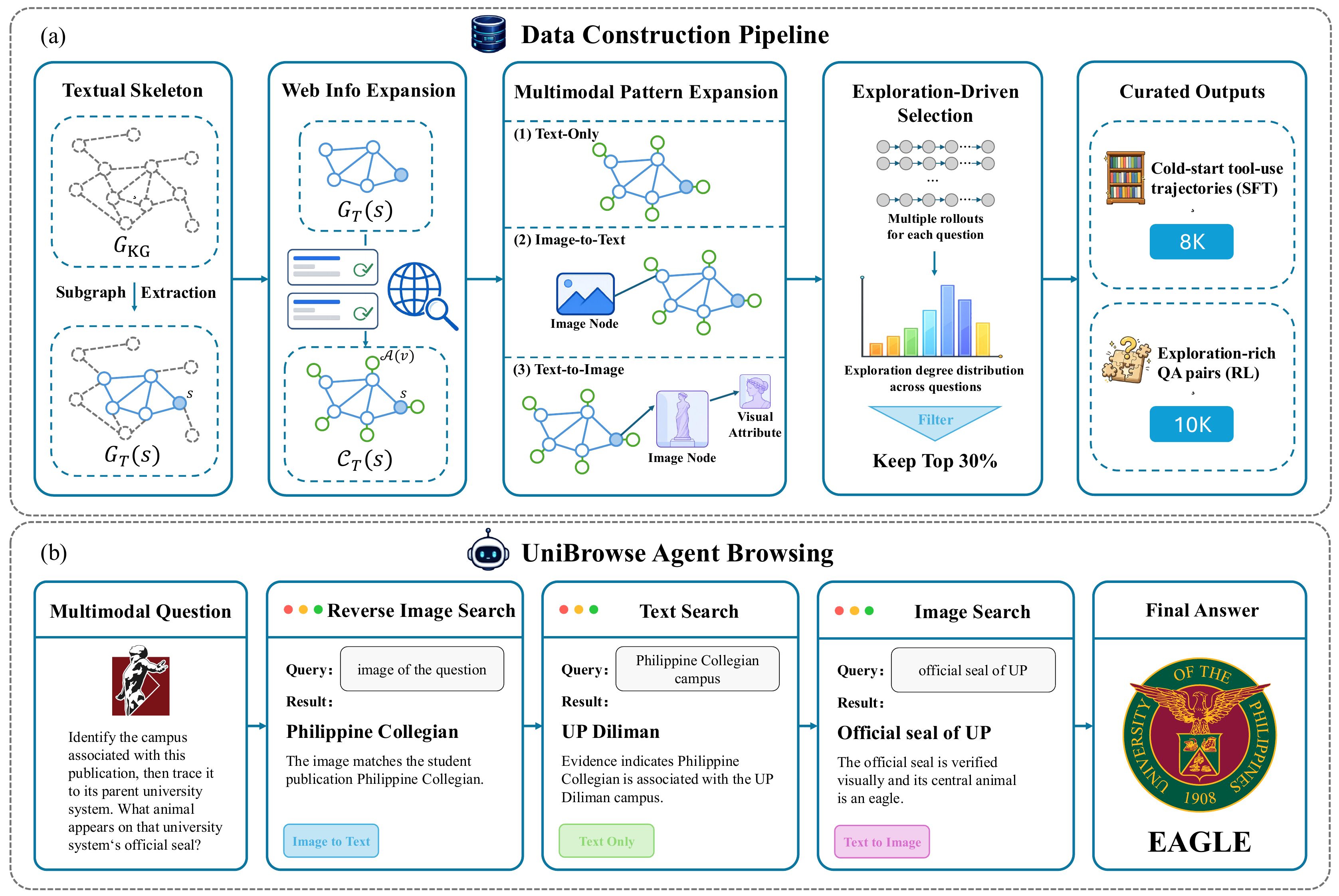}
    \caption{Overview of the \modelname\ framework.
(a) The data construction pipeline starts from textual knowledge-graph skeletons, augments them with live web evidence, expands them into text-only, image-to-text, and text-to-image patterns, and applies exploration-aware selection to RL QA pairs.
(b) The \modelname\ multimodal browsing trajectory shows how the agent resolves an image clue, follows textual web evidence, and verifies the final visual attribute through image search.}
    \label{fig:main}
\end{figure*}

To address these limitations, we propose \textbf{\modelname}, a unified framework comprising a data construction pipeline and a state-of-the-art multimodal browsing agent trained from its output as shown in Figure~\ref{fig:main}. 
\modelname\ expands beyond the previously dominant image-to-text setting by generating multimodal BrowseComp data covering three complementary information-flow patterns (also as pattern): \emph{text-only}, where the answer is obtained by composing textual evidence; \emph{image-to-text}, where an image clue must first be resolved before textual browsing can identify or compose the answer; and \emph{text-to-image}, where textual evidence first identifies what should be searched or verified visually.
Beyond improving pattern coverage, \modelname\ also bridges two previously separate data construction paradigms. It starts from structured knowledge-graph skeletons to preserve controllable reasoning paths, while augmenting them with live web retrieval to introduce realistic, Internet-grounded evidence and reduce the gap between synthetic construction and authentic browsing.
To further improve training efficiency, \modelname\ incorporates exploration-aware data selection into the pipeline. Specifically, we propose the metric of \textit{exploration degree } that measures the diversity of successful reasoning trajectories and filters out low-exploration instances with limited learning signals, enabling reinforcement learning to focus on problems that better exercise long-horizon, multi-step browsing behavior.

From this pipeline, we generate 8K cold-start tool-use trajectories and 10K QA pairs, and train a 35B-scale \modelname\ agent via supervised fine-tuning and exploration-aware RL. The resulting agent achieves state-of-the-art performance on multimodal BrowseComp benchmarks as shown in Figure~\ref{fig:main_results}, validating that our data-to-agent framework effectively translates pattern coverage, web-enhanced grounding, and exploration-driven selection into strong browsing capabilities.

Our contributions can be summarized as follows:
\begin{itemize}[leftmargin=*]
\item We propose \textbf{\modelname}, a unified data construction pipeline that for the first time simultaneously generates multimodal BrowseComp data covering text-only, image-to-text, and text-to-image information-flow patterns, and augments knowledge-graph grounding with live web retrieval for improved real-world fidelity.
\item We introduce an exploration degree metric and an exploration-aware data filtering strategy tailored to the long-horizon nature of BrowseComp tasks, which removes low-exploration instances and significantly improves reinforcement learning efficiency.
\item We train a 35B-scale \textbf{\modelname} agent from 8K cold-start tool-use trajectories and 10K QA pairs produced by our pipeline. The agent achieves state-of-the-art performance on multimodal BrowseComp benchmarks, outperforming prior open-source baselines.
\end{itemize}

\section{Related Work}
\label{sec:related_work}
\subsection{Web Text-Browsing Agents}

Web browsing agents have evolved from browser‑assisted QA~\citep{nakano2021webgpt} and interactive task completion~\citep{yao2022webshop,deng2023mind2web,zhou2023webarena} to complex information seeking over the open web. 
BrowseComp~\citep{wei2025browsecomp} further raises the difficulty: answers must be composed from multiple, non‑obvious web sources, requiring agents to navigate among clues and combine partial evidence. 
To equip agents with such compositional reasoning, recent text‑only BrowseComp pipelines—Tongyi Deepresearch~\citep{team2025tongyi}, WebSailor~\citep{li2025websailor}, WebShaper~\citep{tao2025webshaper}, and OpenSeeker~\citep{du2026openseeker}—have demonstrated that large‑scale synthetic data can significantly improve multi‑step retrieval. 
Despite this progress, these pipelines remain restricted to textual evidence, while real‑world browsing frequently mixes visual and textual clues; constructing multimodal BrowseComp training data therefore remains an open challenge.

\subsection{Multimodal Browsing Agents}
Despite the maturity of text‑only BrowseComp pipelines, extending them to multimodal settings requires agents to jointly reason over images and text in varying orders. 
Recent benchmarks such as MMSearch~\citep{jiang2024mmsearch}, MMSearch+~\citep{tao2025mmsearch}, and FVQA~\citep{wu2025mmsearch} evaluate multimodal search and fact‑based visual QA, while several training pipelines have been proposed: WebWatcher~\citep{geng2025webwatcher} constructs image‑to‑text data from knowledge graphs and introduces BrowseComp‑VL; Vision‑DeepResearch~\citep{zeng2026vision} and Skywork‑R1V4~\citep{zhang2025skywork} use web random walks to generate multimodal trajectories.

However, the data pipelines of these works  cover only the image‑to‑text pattern, lacking the text‑to‑image pattern where textual reasoning identifies a target and visual retrieval provides the final evidence. 
This omission hurts performance on benchmarks covering all three patterns (i.e.,  MM-BrowseComp~\citep{li2025mm}\footnote{MM-BrowseComp is the sole manually curated dataset covering all three patterns and  even leading proprietary workflows fail to reach a 50\% score  on it}). 
Furthermore, while reinforcement learning (RL) can improve long-horizon, multi-turn interaction, prior efforts apply RL directly to unfiltered synthetic data,  wasting computation on low-signal rollouts and limiting performance gains per training step. Together, these gaps motivate the need for training data that covers all information-flow patterns, as well as selection strategies that maximize RL efficiency.

\section{Data Construction Pipeline}
\label{sec:data}

As shown in Figure~\ref{fig:main} (a), we describe the data construction pipeline of \modelname . Given a textual knowledge graph and live web access, the pipeline produces multimodal BrowseComp tasks with three properties critical for training capable agents: (i) full coverage of text-only, image-to-text, and text-to-image information-flow patterns, (ii) web-augmented evidence grounding that bridges structured graph knowledge and open‑web diversity, and (iii) exploration‑aware instance selection that prioritizes problems with rich learning signals for reinforcement learning.


\subsection{Textual KG with Web Expansion} 
\label{sec:text-skeleton}

The pipeline begins with a textual knowledge graph $G_{\mathrm{KG}}$, which provides structured relations among entities and attributes.
To generate a question, for each seed entity $s$ we construct a connected textual evidence subgraph $G_T(s)$ by selecting a set of related anchors and expanding the local neighborhood through multi‑hop graph traversal. This subgraph contains a network of interrelated entities, constraints, and factual edges, and it defines the full evidential context that will later ground the question.

The subgraph $G_T(s)$ is organized into layers that reflect the logical dependencies among evidence nodes.
The initial layer $\mathcal{L}_0$ consists of the seed clues (entities or constraints that are known at the start of the problem).
Each subsequent layer $\mathcal{L}_k$ ($k \ge 1$) consists of evidence nodes that can be determined by aggregating information from preceding layers $\mathcal{L}_0, \dots, \mathcal{L}_{k-1}$.
The final layer $\mathcal{L}_L$ contains only the target answer $a$.
This layered structure ensures that arriving at the answer requires progressively resolving intermediate evidence from the available context.
The total number of intermediate layers controls reasoning depth, and the graph relations guarantee that all required evidential links are present in $G_T(s)$.

\paragraph{Web‑augmented attribute expansion.}
Purely graph‑grounded skeletons may lack the descriptive richness needed for realistic BrowseComp tasks. 
We therefore enrich each entity node $v \in G_T(s)$ with attributes extracted from live web retrieval. 
For each $v$, we issue a web search, collect raw text snippets, and use an LLM to extract candidate attribute statements, retaining only those independently corroborated across multiple search results. 
This yields a verified attribute set $\mathcal{A}(v)$, which is attached to $v$ so that the node carries both its original graph relations and reliable, web‑sourced descriptors. 
Further implementation details, including the extraction prompt, normalization procedure, and support thresholds, are provided in the Appendix~\ref{app:web-attribute-expansion}.

The combined evidence context for $s$ becomes
\begin{equation}
\mathcal{C}_T(s) = \bigl(G_T(s), \{\mathcal{A}(v)\}_{v \in G_T(s)}\bigr),
\end{equation}
which grounds the question in curated graph relations and in verified web attributes.

\subsection{Unified Multimodal Pattern Expansion}
\label{sec:multimodal-expansion}

The text‑only skeleton produces questions that depend solely on textual evidence composition. Real‑world multimodal browsing, however, also requires two additional patterns: image‑to‑text where a visual clue initiates the search, and text‑to‑image where textual reasoning identifies a target whose visual properties supply the final evidence. Rather than constructing these patterns independently, \modelname\ evaluates each generated evidence subgraph $G_T(s)$ for its ability to support multiple information-flow patterns, then materializes the feasible ones by attaching image nodes to the appropriate layers.

\paragraph{Image-to-text.}
Following WebWatcher~\citep{geng2025webwatcher}, we assess whether the entities in the initial layer $\mathcal{L}_0$ have retrievable visual representations. For each candidate entity in $\mathcal{L}_0$, we query image sources and retain the first valid image; entities that pass this check form the image node set $\mathcal{U}_{\mathrm{i2t}}(\mathcal{L}_0)$.

\paragraph{Text-to-image.}
The text‑to‑image pattern imposes stricter requirements on the answer entity $a$ in the final layer $\mathcal{L}_L$. The entity must (i) be \textbf{clearly identifiable} in a retrieved image, so that the agent can reliably recognize it after textual reasoning narrows the candidate set, and (ii) possess a \textbf{distinct, visually verifiable attribute} such as a specific color, shape, logo, or architectural feature, that is \emph{not} stated in the textual evidence. A mere depiction of the target entity is insufficient: the textual steps already identify which entity to look for, so the image must contribute evidence that is not recoverable from text alone.

To determine whether a subgraph can support this pattern, we apply a filtering and annotation stage over the answer entity $a$. We retrieve its candidate images and use a multimodal judge to verify entity identifiability, extract a unique visual attribute absent from the textual evidence, and confirm that removing the image node renders the question unanswerable. Entities whose images pass all three checks form the set $\mathcal{U}_{\mathrm{t2i}}(a)$. Full implementation details of the retrieval, judging, and annotation procedure are provided in the Appendix~\ref{app:multimodal-expansion}.

\paragraph{Pattern materialization.}
From evidence subgraphs ${G_T(s)}$, we instantiate the following patterns:
\begin{itemize}[leftmargin=*]
    \item \textbf{Text-only}: the question is generated directly from the layered evidence structure of $G_T(s)$.
    \item \textbf{Image-to-text}: if $\mathcal{U}_{\mathrm{i2t}}(\mathcal{L}_0) \neq \emptyset$, an image node is linked to the corresponding entity in the initial layer $\mathcal{L}_0$, serving as the entry point of the question.
    \item \textbf{Text-to-image}: if $\mathcal{U}_{\mathrm{t2i}}(a) \neq \emptyset$, an image node carrying the verified visual attribute is appended to the answer entity $a$ in the final layer $\mathcal{L}_L$, providing the final visual evidence for the question.
\end{itemize}
These patterns are not mutually exclusive: all three information-flow patterns can be freely combined within a single question. For instance, a subgraph that supports both image‑to‑text and text‑to‑image conditions can produce a hybrid question carrying both an entry image at $\mathcal{L}_0$ and a verification image at $\mathcal{L}_L$ (see Figure~\ref{fig:case-study} for an illustration), while text‑only evidence composition remains the backbone that connects them. For multimodal patterns, the evidence context is augmented with the same web‑retrieved attributes as in the text‑only case, forming the multimodal context $\mathcal{C}_M(s)$.

\paragraph{Question generation.}
We convert each evidence subgraph (with its attached image nodes, if any) into a natural language question using an VLM generator $\mathrm{Gen}$. For text-only instances, $\mathrm{Gen}$ takes $G_T(s)$ and $\mathcal{C}_T(s)$ as input and produces $(q_T, a^*)$; for multimodal instances, it receives the image-expanded subgraph and $\mathcal{C}_M(s)$, producing $(q_M, a^*)$.

\subsection{Data Filtering and Exploration-Driven Selection}
\label{sec:data-filtering}

Each generated sample---whether text-only, image-to-text, or text-to-image---carries the construction-time key evidence steps derived from its layered subgraph, denoted by $\pi$, and a pattern-specific context $\mathcal{C}_p(s)$ (where $\mathcal{C}_p(s)=\mathcal{C}_T(s)$ for text-only instances and $\mathcal{C}_p(s)=\mathcal{C}_M(s)$ for multimodal ones). Before being used for training, every sample first passes a set of lightweight static validations. We then introduce an exploration degree metric that is applied \emph{only} to the reinforcement learning stage, where training cost is highest.

\paragraph{Basic quality control.}
We enforce three fast checks:
\begin{enumerate}[leftmargin=*,nosep]
    \item \textbf{Sufficiency} – the answer $a^*$ must be fully supported by $\mathcal{C}_p(s)$.
    \item \textbf{Modality necessity} – for image-involving patterns, removing the image node must render the question unanswerable, guaranteeing that the visual clue is non-redundant.
    \item \textbf{Uniqueness} – no alternative answer may satisfy all constraints in the evidence context.
\end{enumerate}
Samples that fail any check are discarded.

\paragraph{Exploration degree.}
Static validation ensures well-formedness but not whether a problem exposes a rich search space for policy learning.
Informative RL instances should require meaningful decisions---query reformulation, source selection, evidence verification, modality switching---rather than admit only a single narrow solution path.
When all successful rollouts follow essentially the same route, they provide little behavioral contrast for the policy to learn from.
We capture this diversity through the \textbf{exploration degree} $\mathcal{E}(q)$.
For a validated question $q$ with its key evidence steps $\pi$, we sample $B$ rollout trajectories using a base agent and let $\mathcal{Y}^{+}(q)$ be the subset that reaches the correct answer.
If $|\mathcal{Y}^{+}(q)| \le 1$, we set $\mathcal{E}(q)=0$.
Otherwise, for each successful trajectory $\tau_b$ we measure its exploration length relative to the fixed key-evidence-step set, i.e., the number of browsing actions (search / linksummary calls) it takes, normalized by $|\pi|$:
\begin{equation}
\ell(\tau_b, \pi) = \frac{\text{\# browsing steps in } \tau_b}{|\pi|},
\end{equation}
and take the variance of this normalized length across successful rollouts:
\begin{equation}
\mathcal{E}(q) = \mathrm{Var}_{\tau_b \in \mathcal{Y}^{+}(q)}\big[\ell(\tau_b, \pi)\big].
\end{equation}
Since $|\pi|$ is fixed per question, $\ell$ isolates how much the browsing effort of successful routes varies. A high $\mathcal{E}(q)$ indicates that correct answers are reached through browsing routes of substantially different lengths---some via a compact direct path, others via reformulation-heavy or backtracking exploration---exposing the policy to a broader but grounded exploration space and providing richer contrastive signals for RL.
A low $\mathcal{E}(q)$ signals that all successful routes are essentially uniform in length, offering limited behavioral contrast.
We use $\mathcal{E}(q)$ exclusively for data selection, not as an auxiliary reward; this metric guides instance filtering in subsequent RL training.
We provide a further detailed explanation and a worked example in Appendix~\ref{app:exploration-degree-example}.
\section{Training \modelname\ Agent}
\label{sec:training}

We convert each QA pair from Section~\ref{sec:data} into a tool‑augmented trajectory and train the agent in two stages: supervised fine‑tuning (SFT) establishes a stable action grammar, while reinforcement learning (RL) on exploration‑rich instances optimizes multi‑step browsing policies, resulting the \modelname\ Agent as shown in Figure~\ref{fig:main} (b).

\subsection{Tool Environment and Trajectory Format}
\label{sec:tool_trajectory}

The agent interacts with two tools that mirror the evidence‑gathering patterns defined in our data pipeline.
\begin{itemize}[leftmargin=*,nosep]
    \item \textsc{Search} (backed by SerpAPI) supports three modes: \texttt{text\_only} for standard web retrieval, \texttt{image\_to\_text} for resolving an image clue into textual evidence, and \texttt{text\_to\_image} for locating visual evidence from a textual description.
    \item \textsc{LinkSummary} takes a URL and a query; an auxiliary Qwen3‑30B‑A3B model~\citep{yang2025qwen3} reads the page and returns a query‑conditioned summary, keeping raw HTML outside the main agent’s context.
\end{itemize}

Browsing is formulated as a ReAct‑style sequential decision process~\citep{yao2022react}. At step $t$, the agent observes state $S_t=(q,p,h_t)$ (question, pattern, serialized history) and produces an action $z_t\sim\pi_\theta(\cdot|S_t)$. Actions can be reasoning steps, tool calls, or a final answer. Tool observations $o_t$ are appended to the history, and a trajectory terminates when the agent emits an answer segment.

All actions and observations follow a fixed serialization protocol, summarized in Table~\ref{tab:trajectory_format}.

\begin{table}[t]
\centering
\footnotesize
\begin{trajectorybox}
\setlength\tabcolsep{3pt}
\renewcommand{\arraystretch}{1.15}
\begin{tabularx}{\linewidth}{@{}p{0.25\linewidth} X@{}}
\textbf{Segment} & \textbf{Serialized template} \\
\midrule
\tagrow{Reasoning}{\thinktag\ $r_t$ \tagclose{tagthink}{think}}
\addlinespace[2pt]
\tagrow{Tool call}{\toolcalltag\ $c_t$ \tagclose{tagcall}{tool\_call}}
\addlinespace[2pt]
\tagrow{Tool return}{\toolresponsetag\ $o_t$ \tagclose{tagresponse}{tool\_response}}
\addlinespace[2pt]
\tagrow{Final answer}{\answertag\ $\hat{a}$ \tagclose{taganswer}{answer}}
\end{tabularx}
\end{trajectorybox}
\caption{Typed serialization format for \modelname\ tool‑augmented trajectories.}
\label{tab:trajectory_format}
\end{table}

\subsection{Training Overview}
\label{sec:training_overview}

We train \modelname\ in two stages, using the data produced by the pipeline of Section~\ref{sec:data}.
\paragraph{Cold-start SFT.} We use questions produced by the pipeline to produce approximately 8,000 tool‑use trajectories, filtered only by basic quality checks; this instills a stable action grammar and a pattern‑aware prior.
\paragraph{Exploration‑aware RL.} To improve robustness, we further optimize the SFT model with reinforcement learning. For this expensive stage, we activate exploration‑aware selection (Section~\ref{sec:data-filtering}), retaining the top‑$k$ quantile instances by exploration degree $\mathcal{E}(q)$ to obtain around 10,000 exploration‑rich problems. The RL reward combines the correctness of the answers with the correctness of the format, and we optimize using Group Relative Policy Optimization (GRPO)~\citep{shao2024deepseekmath}. The set of key evidence steps $\pi$ recorded during data construction is used only for data selection and analysis, not in the reward itself.
Full details of teacher trajectory construction, loss functions, reward design, and hyperparameters are provided in the Appendix~\ref{app:multimodal-expansion}.

\section{Experiments}

In this section, we evaluate \modelname\ on multimodal browsing tasks. We first describe the experimental setup, including implementation details, benchmarks, baselines, and metrics. We then compare \modelname\ with a broad set of baselines. Finally, we conduct ablation studies to validate our key design choices.

\input{tables/main_results}

\subsection{Experimental Setup}

\paragraph{Implementation Details}
We implement \modelname\ training on 64 NVIDIA H20 GPUs using veRL~\citep{wu2025hybridflow} and initialize the agent from Qwen3.5-35B-A3B~\citep{qwen35blog}. The training data is produced by the \modelname\ construction and validation pipeline described in Section~\ref{sec:data}; unless otherwise specified, we balance the three information-flow patterns (text-only, image-to-text, and text-to-image) with an equal 1:1:1 proportion. The SFT and RL configurations follow the setup described in Section~\ref{sec:training_overview}, with full training details provided in Appendix~\ref{app:training}.


\paragraph{Benchmarks}
We evaluate on five benchmarks. BrowseComp-VL (BC-VL) is introduced by WebWatcher~\citep{geng2025webwatcher}. MM-BrowseComp (MM-BC) is a multimodal BrowseComp benchmark~\citep{li2025mm} and, crucially, the only benchmark that evaluates all three information-flow patterns (text-only, image-to-text, and text-to-image), making it the most challenging in our suite. For MM-BC, we use the subset that does not require video reasoning or map-specific tools, so that all models are compared under the same image-and-web browsing tool setting. MMSearch+ is evaluated on its single-image subset~\citep{tao2025mmsearch}, while MMSearch follows the original multimodal search benchmark~\citep{jiang2024mmsearch}. FVQA is the fact-based visual question answering benchmark introduced in MMSearch-R1~\citep{wu2025mmsearch}.

\paragraph{Baselines and Metrics}
We compare \modelname\ with three groups of baselines. Direct-answer MLLMs answer the question without an explicit browsing workflow, measuring how much the base model can solve from its parametric and visual reasoning ability alone. Agent-workflow MLLMs use tool-augmented browsing trajectories, providing a stronger comparison for search-and-reasoning settings. DeepResearch MLLMs are recent open-source systems specialized for search-intensive reasoning. For our own model family, we report the Qwen3.5-35B-A3B base model, the \modelname\ SFT model, and the \modelname\ SFT+RL model to isolate the effect of cold-start trajectory learning and reinforcement learning. We report LLM-as-a-judge accuracy on each benchmark.

\subsection{Main Results}

Table~\ref{tab:main_results} compares proprietary and open-source systems under three model settings.
Direct answering without tool use proves largely insufficient: among models with complete results, the best direct-answer average is only 38.6\% (Gemini-3.1 Pro), confirming that internal knowledge alone cannot reliably solve these multimodal BrowseComp tasks. 
Equipping models with browsing tools yields dramatic improvements across the board. 
The strongest proprietary agent workflow, Gemini-3.1 Pro, reaches an average of 63.2\%.
This underscores that tool-augmented interaction is essential for compositional multimodal retrieval.

Our \modelname\ agent, trained with the proposed data pipeline, establishes a new state of the art among open-source models. 
Starting from the Qwen3.5-35B-A3B baseline (43.9\% average), supervised fine-tuning on our cold-start trajectories lifts the average to 50.6\%, already surpassing all prior open-source multimodal browsing agents with reported averages (e.g., WebWatcher-32B at 31.7\%). 
Adding exploration-aware reinforcement learning further boosts performance to 54.4\% average, a gain of 10.5 percentage points over the base model, achieving the best results on every benchmark in the open-source category: 57.8 on BrowseComp-VL, 28.2 on MM-BrowseComp, 34.1 on MMSearch+, 75.2 on MMSearch, and 76.7 on FVQA.
While a gap remains to the strongest proprietary workflow (63.2\%), \modelname\ narrows it by more than half.
These results demonstrate that a unified data generation strategy covering diverse information-flow patterns, combined with exploration-aware data selection, can produce highly capable multimodal browsing agents from open-source foundations. Additionally, we provide a case study section in the Appendix~\ref{app:case-study}.

\subsection{Ablation Study}

\paragraph{Data ablation.}
Table~\ref{tab:ablation_results} summarizes a data ablation on BC-VL and MM-BC. The variants use the same training recipe but differ in which patterns are included in the training data.

\input{tables/ablation_results}

Training only on text-only data drops below the base model, from 33.8 to 30.4 Avg., showing that post-training without multimodal tool-use patterns can hurt browsing performance. Adding image-to-text data recovers much of this loss and improves BC-VL from 46.6 to 54.3, confirming the value of image-grounded search supervision. However, MM-BC remains at the base level. This limited gain on MM-BC is consistent with its mixed information-flow composition, where image-to-text supervision alone does not cover text-to-image cases. The full mixture, which further adds text-to-image data, improves MM-BC to 28.2 and reaches 43.0 Avg., indicating that broad pattern coverage is important for multimodal BrowseComp training.

\paragraph{Exploration-selection ablation.}
Table~\ref{tab:selection_ablation_results} studies the quantile threshold used to select RL questions by exploration degree. All variants use the same number of RL questions and differ only in the retained candidate pool before sampling.

Keeping all candidates or using a loose threshold yields limited gains, suggesting that many low-contrast questions provide weak RL signal. Performance generally improves when overly broad pools are narrowed toward higher-exploration questions, peaking at the top-30\% threshold. The top-10\% setting drops, suggesting that overly strict filtering reduces data diversity. We therefore use the top 30\% threshold in the final \modelname\ agent.

\section{Conclusion}
We presented \modelname, a data-to-agent framework for multimodal BrowseComp.
\modelname\ constructs training data across the full set of information-flow patterns: text-only, image-to-text, and text-to-image.
It further grounds graph-structured reasoning paths with live web evidence, combining the controllability of knowledge-graph-based construction with the diversity and realism of open-web retrieval.
To make post-training more efficient, \modelname\ introduces exploration degree and uses it for exploration-aware RL data selection, focusing expensive rollout-based optimization on questions that expose richer browsing behaviors.
From the resulting cold-start tool-use trajectories and exploration-rich QA pairs, we train a 35B-A3B multimodal browsing agent that achieves strong open-source performance across five multimodal browsing benchmarks.
These results show that controllable multimodal data construction and exploration-aware post-training are both important for building effective browsing agents. 

\section*{Limitations}
Although \modelname\ substantially narrows the gap between open-source agents and the strongest proprietary workflow, a performance gap still remains, suggesting that stronger foundation models, larger-scale post-training, and more capable tool infrastructures may further improve multimodal browsing agents. Our experiments are limited to public-web, image-and-text browsing tasks that can be handled with search and link summarization tools; richer environments such as interactive webpages, video- or map-centric browsing, multilingual search, and domain-specific evidence sources remain beyond the capabilities of the current agent. On the reinforcement learning side, while our exploration-aware selection provides an effective principle for filtering training data, we do not address the long-tail inefficiencies inherent in agentic rollout trajectories, which remain a major bottleneck for RL training efficiency. Similarly, our exploration-aware selection relies on a fixed rollout budget to estimate the exploration degree; the design of more sample-efficient selection strategies that adapt to evolving web content and model capabilities also remains an open challenge.

\section*{Ethics Statement}
This work fully complies with the ACL Ethics Policy.
We declare that there are no ethical issues in this paper, to the best of our knowledge.

\bibliography{bib/custom}
\bibliographystyle{bib/acl_natbib}

\input{appendix}

\clearpage

\end{document}

%% file: math_commands.tex
\usepackage{amsmath,amsfonts,bm}

\def\eqref#1{equation~\ref{#1}}

\def\1{\bm{1}}

\DeclareMathAlphabet{\mathsfit}{\encodingdefault}{\sfdefault}{m}{sl}
\SetMathAlphabet{\mathsfit}{bold}{\encodingdefault}{\sfdefault}{bx}{n}



%% file: tables/main_results.tex
\begin{table*}[t]
\vspace{-15mm}
\centering
\footnotesize
\setlength{\tabcolsep}{5pt}
\renewcommand{\arraystretch}{1.12}
\arrayrulecolor{tableline}
\begin{adjustbox}{max width=\textwidth}
\begin{tabular}{lcccccc}
\toprule[1.2pt]
\rowcolor{tablehead}
\textbf{Model} &
\textbf{BC-VL} &
\textbf{MM-BC} &
\textbf{MMSearch+} &
\textbf{MMSearch} &
\textbf{FVQA} &
\textbf{Avg.} \\
\midrule
\rowcolor{tablesection}
\multicolumn{7}{c}{\textbf{Direct Answer MLLMs}} \\
\midrule
GPT-5 & 47.2 & 7.9 & 19.1 & 33.3 & 57.3 & 33.0 \\
Gemini-2.5 Pro & 43.1 & 7.9 & 14.5 & 39.8 & 60.7 & 33.2 \\
Gemini-2.5 Flash & 37.1 & 4.8 & 8.1 & 30.4 & 47.7 & 25.6 \\
Gemini-3.1 Pro & 42.8 & \textbf{16.7} & \textbf{21.6} & \textbf{55.6} & 56.2 & \textbf{38.6}\\ 
Claude-4-Sonnet & 29.3 & -- & 4.0 & 18.7 & 35.3 & -- \\
Claude-3.7-Sonnet & 32.3 & -- & 4.0 & 21.1 & 36.7 & -- \\
Doubao Seed 2.0 Pro & \textbf{47.9} & 9.72 & 14.9 & 45.6 & \textbf{62.2} & 36.1 \\
Kimi K2.5~\citep{team2026kimi} & 40.9 & 9.72 & 11.7 & 58.5 & 60.3 & 36.2 \\
Qwen3-VL-30B-A3B-Thinking~\citep{bai2025qwen3} & 34.6 & 5.2 & 4.5 & 19.3 & 32.7 & 19.3 \\
Qwen3.5-35B-A3B~\citep{qwen35blog} & 31.1 & 8.30 & 0.27 & 28.7 & 39.6 & 21.6 \\
Qwen3.5-122B-A10B~\citep{qwen35blog} & 35.4 & 9.80 & 5.11 & 38.9 & 45.2 & 26.9 \\

\midrule
\rowcolor{tablesection}
\multicolumn{7}{c}{\textbf{Agent Workflow MLLMs}} \\
\midrule
GPT-5 & 46.1 & 18.5 & 17.2 & 63.7 & 69.0 & 42.9 \\
Gemini-2.5 Pro & 49.9 & 14.8 & 22.2 & 69.0 & 68.3 & 44.8 \\
Gemini-2.5 Flash & 44.6 & 10.2 & 19.9 & 64.0 & 68.0 & 41.3 \\
Gemini-3.1 Pro & 64.7 & \textbf{46.3} & \textbf{49.7} & 77.0 & \textbf{78.3} & \textbf{63.2}\\
Claude-4-Sonnet & 48.6 & -- & 23.1 & 67.2 & 69.0 & -- \\
Claude-3.7-Sonnet & 50.4 & -- & 17.2 & 63.7 & 67.3 & -- \\
Doubao Seed 2.0 Pro & \textbf{67.3} & 38.4 & 49.1 & \textbf{80.1} & 77.5 & 62.5 \\
Kimi K2.5~\citep{team2026kimi} & 52.8 & 32.4 & 35.9 & 64.9 & 67.9 & 50.8 \\
Qwen3-VL-30B-A3B-Thinking~\citep{bai2025qwen3} & 44.1 & 6.00 & 13.6 & 53.2 & 63.0 & 36.0 \\
Qwen3.5-35B-A3B~\citep{qwen35blog} & 50.4 & 17.1 & 24.4 & 62.6 & 65.2 & 43.9 \\
Qwen3.5-122B-A10B~\citep{qwen35blog} & 53.0 & 25.9 & 32.9 & 65.1 & 71.0 & 49.6 \\

\midrule
\rowcolor{tablesection}
\multicolumn{7}{c}{\textbf{Multimodal DeepResearch MLLM}} \\
\midrule
MMSearch-R1-7B & -- & -- & -- & 53.8 & 58.4 & -- \\
WebWatcher-7B & 20.3 & -- & -- & 49.1 & -- & -- \\
WebWatcher-32B & 26.7 & 10.7 & 13.6 & 55.3 & 52.1 & \textbf{31.7} \\
Skywork-R1V4-30B-A3B~\citep{zhang2025skywork} & 38.4 & -- & -- & 66.1 & 67.2 & -- \\
Vision-DeepResearch-8B & 42.6 & -- & 20.4 & \textbf{69.6} & 64.7 & -- \\
Vision-DeepResearch-30B-A3B & \textbf{53.7} & -- & \textbf{28.5} & \textbf{69.6} & \textbf{74.2} & -- \\

\midrule
\rowcolor{ourssection}
\multicolumn{7}{c}{\textbf{UniBrowse Agent (Ours)}} \\
\midrule
\rowcolor{oursbase}
Qwen3.5-35B-A3B~\citep{qwen35blog} & 50.4 & 17.1 & 24.4 & 62.6 & 65.2 & 43.9 \\
\rowcolor{ourssft}
UniBrowse-35B-A3B SFT & 54.4 & 21.8 & 32.6 & 70.8 & 73.4 & 50.6 \\
\rowcolor{oursrl}
UniBrowse-35B-A3B SFT+RL & \textbf{57.8} & \textbf{28.2} & \textbf{34.1} & \textbf{75.2} & \textbf{76.7} & \textbf{54.4} \\
\bottomrule[1.2pt]
\end{tabular}
\end{adjustbox}
\arrayrulecolor{black}
\captionsetup{hypcap=false}
\caption{Main results on multimodal browsing reasoning benchmarks. We report accuracy on each benchmark; Avg. is shown only when all five benchmark results are available. The best result in each block is highlighted in \textbf{bold}.}
\label{tab:main_results}
\end{table*}

%% file: tables/ablation_results.tex
\begin{table}[t]
\centering
\footnotesize
\setlength{\tabcolsep}{4pt}
\renewcommand{\arraystretch}{1.12}
\begin{tabularx}{\columnwidth}{@{}>{\raggedright\arraybackslash}X c c c@{}}
\toprule
\textbf{Training data} & \textbf{BC-VL} & \textbf{MM-BC} & \textbf{Avg.} \\
\midrule
Base model & 50.4 & 17.1 & 33.8 \\
\midrule
Text-only & 46.6 & 14.1 & 30.4 \\
Text-only + image-to-text & 54.3 & 18.5 & 36.4 \\
\textbf{All three patterns} & \textbf{57.8} & \textbf{28.2} & \textbf{43.0} \\
\bottomrule
\end{tabularx}
\caption{Data ablation on BrowseComp-VL (BC-VL) and MM-BrowseComp (MM-BC). Avg. is the mean of the two benchmark scores. ``All three patterns'' includes text-only, image-to-text, and text-to-image data.}
\label{tab:ablation_results}
\end{table}

\begin{table}[t]
\centering
\footnotesize
\setlength{\tabcolsep}{4pt}
\renewcommand{\arraystretch}{1.12}
\begin{tabularx}{\columnwidth}{@{}>{\raggedright\arraybackslash}X c c c@{}}
\toprule
\textbf{RL selection pool} & \textbf{BC-VL} & \textbf{MM-BC} & \textbf{Avg.} \\
\midrule
Top 100\% by $\mathcal{E}(q)$ & 53.9 & 24.7 & 39.3 \\
Top 70\% by $\mathcal{E}(q)$ & 55.1 & 23.3 & 39.2 \\
Top 50\% by $\mathcal{E}(q)$ & 56.5 & 26.7 & 41.6 \\
\textbf{Top 30\% by $\mathcal{E}(q)$} & \textbf{57.8} & \textbf{28.2} & \textbf{43.0} \\
Top 10\% by $\mathcal{E}(q)$ & 52.1 & 20.3 & 36.2 \\
\bottomrule
\end{tabularx}
\caption{Exploration-selection ablation. Each row uses 10K RL questions sampled from the retained pool ranked by $\mathcal{E}(q)$.}
\label{tab:selection_ablation_results}
\end{table}

%% file: appendix.tex
\clearpage

\appendix

\section{Case Study}
\label{app:case-study}

\begin{figure*}[t]
\centering
\vspace{-15mm}
\includegraphics[width=1.0\textwidth]{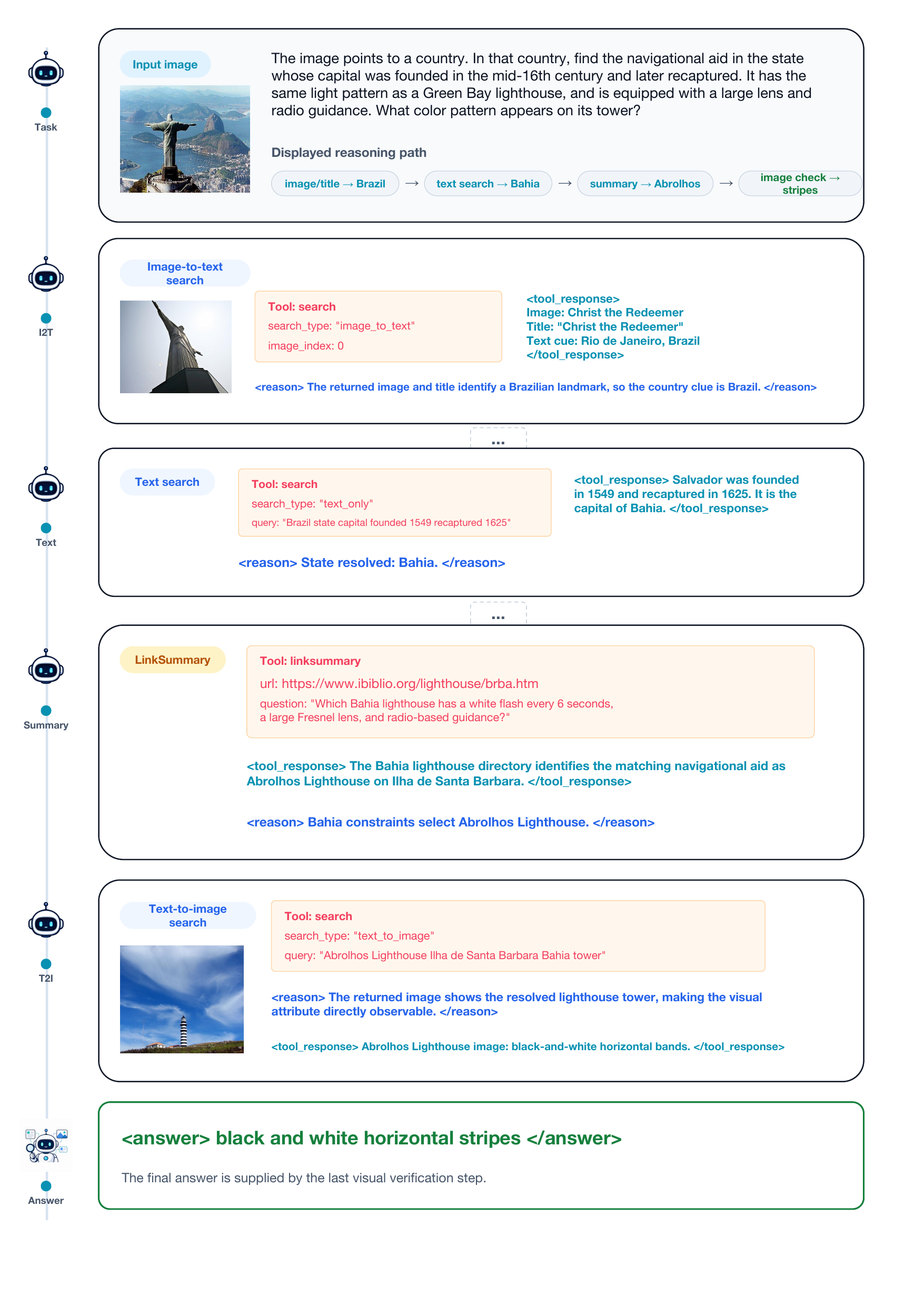}
\vspace{-15mm}
\caption{A case study of \modelname\ on a multimodal browsing question.}
\label{fig:case-study}
\end{figure*}

Figure~\ref{fig:case-study} shows a concrete text-to-image pattern question produced by \modelname\ data pipeline. In this example, the \modelname\ agent first grounds the image clue, uses textual search and link summarization to resolve the target entity, and finally performs text-to-image search to verify the visual attribute required by the answer.

\section{Multimodal Pattern Expansion Details}
\label{app:multimodal-expansion}

This appendix supplements Section~\ref{sec:multimodal-expansion} with implementation details of the image retrieval, multimodal judging, and annotation steps for constructing $\mathcal{U}_{\mathrm{i2t}}(\mathcal{L}_0)$ and $\mathcal{U}_{\mathrm{t2i}}(a)$.

\paragraph{Image-to-text image set.}
For the image-to-text pattern, we follow the procedure of WebWatcher~\citep{geng2025webwatcher}. 
Given the entities in the initial layer $\mathcal{L}_0$, we query Google Images with each entity name and its graph-provided aliases, retrieve the top-10 results, and retain the first image that is not a generic icon or logo unrelated to the entity. 
The retained image becomes the sole candidate in $\mathcal{U}_{\mathrm{i2t}}(\mathcal{L}_0)$.

\paragraph{Text-to-image candidate retrieval.}
For each answer entity $a$ in the final layer $\mathcal{L}_L$, we construct a set of search queries by combining the entity name with contextual keywords from the layered subgraph (e.g., attributes of entities in intermediate layers $\mathcal{L}_1, \dots, \mathcal{L}_{L-1}$). 
We issue these queries to Google Images and collect up to 20 candidate images per query, deduplicating near-duplicates via perceptual hashing. 
All unique images form the initial candidate pool for $a$.

\paragraph{Multimodal judge.}
We use an MLLM (the same judge used for quality control in Section~\ref{sec:data-filtering}) to filter and annotate text-to-image candidates. 
The judge receives the question context (the layered evidence structure of the subgraph together with the final answer $a$) together with a candidate image. 
It is asked to answer three structured questions:
\begin{enumerate}[leftmargin=*,nosep]
    \item \textbf{Entity identifiability:} Is the target entity $a$ clearly depicted and distinguishable in the image? (yes/no)
    \item \textbf{Visual attribute extraction:} What visually verifiable attribute of the entity can be observed in the image that is \emph{not} already stated in the layered textual evidence? If none, answer ``None''.
    \item \textbf{Modality necessity:} If the image were removed, would the question still be answerable from text alone? (yes/no)
\end{enumerate}
A candidate image is retained only if the judge answers ``yes'' to question 1, provides a non-empty, specific attribute for question 2, and answers ``no'' to question 3. 
The extracted attribute is then normalized to a short phrase (e.g., ``red roof'', ``golden statue'', ``three arches'').

\paragraph{Final image set.}
For each answer entity $a$, we keep the first three images that pass the judge filter to form $\mathcal{U}_{\mathrm{t2i}}(a)$. 
If fewer than three images pass, we discard the entity from text-to-image expansion (it will only appear in text-only and, if applicable, image-to-text patterns). 
All retained images are stored together with their verified visual attribute for use in the pattern expansion step.

\subsection{Cold‑Start Supervised Fine‑Tuning}
\label{sec:cold_start_sft}

To obtain reliable teacher demonstrations, we prompt an open‑source multimodal teacher model (Kimi K2.5~\citep{team2026kimi}) with each validated task $x=(q,\mathcal{C}_p(s),a^*,p)$ and collect rollouts that reach the correct answer while respecting the tool protocol and pattern constraints. The resulting teacher trajectory $\tau^\star=(z_1,o_1,\dots,z_T,\hat{a})$, with $\hat{a}=a^*$, provides a step‑by‑step demonstration of correct tool use and evidence composition.

The cold‑start SFT set is $\mathcal{D}_{\mathrm{cs}}=\{(x_i,\tau_i^\star)\}_{i=1}^{N_{\mathrm{cs}}}$, and we minimize the standard next‑token prediction loss:
\begin{equation}
\mathcal{L}_{\mathrm{cs}}(\theta)= -\sum_{(x,\tau^\star)\in\mathcal{D}_{\mathrm{cs}}} \sum_{z_t\in\tau^\star}\log \pi_\theta(z_t\mid S_t).
\end{equation}
We train for 3 epochs and use the final checkpoint. This stage establishes a stable action grammar and a pattern‑aware tool‑use prior, but it does not directly penalize inefficient exploration or reward recovery from mistakes.

\subsection{Reinforcement Learning with Exploration‑Aware Data Selection}
\label{sec:rl_training}

Cold‑start SFT gives the model a basic browsing protocol, but robust deployment requires the ability to choose effective strategies and avoid degenerate loops. We therefore further optimize the policy through reinforcement learning. Crucially, RL is applied only to a subset of validated QA pairs selected via the exploration degree introduced in Section~\ref{sec:data-filtering}.

For cold-start supervised fine-tuning, we do \emph{not} apply exploration-based filtering; we keep all teacher-generated trajectories that pass basic quality control to establish a stable action grammar and pattern‑aware prior. For reinforcement learning, where training resource consumption is greatest, we activate exploration-aware selection: from the pool of validated QA pairs, we retain only those instances whose exploration degree $\mathcal{E}(q)$ exceeds a top‑$k$ quantile threshold. This yields a compact set of exploration-rich problems that maximize policy improvement per training step and prevent wasted computation on low-signal trajectories. Each retained instance carries its set of key evidence steps $\pi$, recorded during construction from the layered evidence subgraph. Together with the QA pair, these form the RL training set $\mathcal{D}_{\mathrm{RL}}$, but $\pi$ is used only for data selection and analysis; the RL reward itself does not use $\pi$ directly. We train the policy for 150 steps using GRPO.

\paragraph{Reward Design.}
We use a composite reward combining answer correctness and format adherence:
\begin{equation}
R(x,\tau)=\lambda_{\mathrm{ans}}R_{\mathrm{ans}}(x,\tau)+\lambda_{\mathrm{fmt}}R_{\mathrm{fmt}}(\tau).
\end{equation}
$R_{\mathrm{ans}}$ employs an LLM‑as‑a‑judge to verify semantic equivalence between the extracted answer and $a^*$. $R_{\mathrm{fmt}}$ checks that the trajectory strictly follows the serialization protocol (correct tag usage, valid tool call structure, and exactly one final answer segment). This design preserves answer accuracy while discouraging format drift and redundant tool chains.

\paragraph{Policy Optimization.}
We adopt Group Relative Policy Optimization (GRPO)~\citep{shao2024deepseekmath}. For each task $x\in\mathcal{D}_{\mathrm{RL}}$, the current policy $\pi_\theta$ samples $K$ trajectories, and the advantage is computed from the reward distribution within the group. The objective is
\begin{equation}
\max_{\theta}\; \mathbb{E}_{x\sim\mathcal{D}_{\mathrm{RL}},\,\tau\sim\pi_\theta(\cdot|x)}\left[R(x,\tau)\right],
\end{equation}
optimized via the standard GRPO update. After RL training, the \modelname\ agent achieves robust multimodal browsing capabilities with improved trajectory efficiency and answer accuracy.

\paragraph{Reward Design.}
We use a composite reward combining answer correctness and format adherence:
\begin{equation}
R(x,\tau)=\lambda_{\mathrm{ans}}R_{\mathrm{ans}}(x,\tau)+\lambda_{\mathrm{fmt}}R_{\mathrm{fmt}}(\tau).
\end{equation}
$R_{\mathrm{ans}}$ employs an LLM‑as‑a‑judge to verify semantic equivalence between the extracted answer and $a^*$. $R_{\mathrm{fmt}}$ checks that the trajectory strictly follows the serialization protocol (correct tag usage, valid tool call structure, and exactly one final answer segment). This design preserves answer accuracy while discouraging format drift and redundant tool chains.

\paragraph{Policy Optimization.}
We adopt Group Relative Policy Optimization (GRPO)~\citep{shao2024deepseekmath}. For each task $x\in\mathcal{D}_{\mathrm{RL}}$, the current policy $\pi_\theta$ samples $K$ trajectories, and the advantage is computed from the reward distribution within the group. The objective is
\begin{equation}
\max_{\theta}\; \mathbb{E}_{x\sim\mathcal{D}_{\mathrm{RL}},\,\tau\sim\pi_\theta(\cdot|x)}\left[R(x,\tau)\right],
\end{equation}
optimized via the standard GRPO update. After RL training, the \modelname\ agent achieves robust multimodal browsing capabilities with improved trajectory efficiency and answer accuracy.

\section{Web-Augmented Attribute Expansion}
\label{app:web-attribute-expansion}

This appendix provides additional implementation details for the web-augmented grounding step in Section~\ref{sec:text-skeleton}. The goal is to enrich each graph entity with attributes that are likely to appear in real web evidence, while avoiding one-off or unsupported claims. For an entity node $v$, \modelname\ searches the web for entity-related pages, extracts candidate attribute statements from each page, and retains only attributes that are repeatedly supported by independent pages.

\begin{algorithm}[H]
\caption{Web-Augmented Attribute Expansion for an Entity Node}
\label{alg:web-attribute-expansion}
\begin{algorithmic}[1]
\Require Entity node $v$, search engine $\mathrm{Search}$, page summarizer $\mathrm{Summarize}$, LLM extractor $\mathrm{Extract}$, minimum page count $N_{\min}=5$, support threshold $\gamma=3$
\Ensure Verified attribute set $\mathcal{A}(v)$
\State Construct search queries $\mathcal{Q}(v)$ using the entity name, aliases, and graph-neighbor hints.
\State Retrieve candidate webpages $\mathcal{P}(v)\leftarrow \mathrm{Search}(\mathcal{Q}(v))$.
\State Initialize candidate multiset $\mathcal{M}\leftarrow \emptyset$.
\ForAll{webpage $p\in \mathcal{P}(v)$}
    \State Obtain a query-conditioned summary $d_p\leftarrow \mathrm{Summarize}(p,v)$.
    \State Extract candidate attributes $\mathcal{C}_p\leftarrow \mathrm{Extract}(v,d_p)$ using the prompt in Figure~\ref{fig:web-attr-extract-prompt}.
    \ForAll{candidate attribute $c\in \mathcal{C}_p$}
        \If{$c$ is specific, entity-grounded, and not contradicted by $d_p$}
            \State Normalize $c$ into a canonical attribute form and add $(c,p)$ to $\mathcal{M}$.
        \EndIf
    \EndFor
\EndFor
\State Group semantically equivalent candidates in $\mathcal{M}$ across webpages.
\State Retain an attribute $c$ if it is supported by at least $\gamma$ distinct webpages and the total number of usable webpages is at least $N_{\min}$.
\State \Return $\mathcal{A}(v)=\{c: \mathrm{support}(c)\geq \gamma\}$.
\end{algorithmic}
\end{algorithm}

\paragraph{Extraction and aggregation.}
In practice, we use at least five usable webpages for each entity whenever available. Each webpage is first summarized with the target entity in mind; the extractor then proposes attributes that can be used as additional descriptors or constraints in question construction. We normalize near-duplicate attributes with an LLM judge, merge semantically equivalent claims, and keep only attributes that appear in at least three independent webpages. This threshold favors stable properties such as roles, affiliations, locations, creators, functions, awards, visible characteristics, or notable associated works, while discarding rare claims that are likely to be noisy, promotional, or context-specific.

\clearpage
\onecolumn
\begin{figure}[H]
\centering
\begin{promptbox}[Per-Page Attribute Extraction Prompt]{promptgreen}
You are helping construct reliable web-grounded attributes for a BrowseComp-style question.\\
\textbf{Entity:} \{entity\_name\}\\
\textbf{Known graph context:} \{neighbor\_entities\_and\_relations\}\\
\textbf{Webpage summary:} \{page\_summary\}\\[2pt]
Extract candidate attributes of the entity that are explicitly supported by the webpage. Each attribute should be useful as a clue, constraint, or descriptor in a multi-hop browsing question. Prefer stable and verifiable properties, such as affiliation, location, creator, time, award, function, material, visual feature, associated work, or role. Do not infer attributes that are not stated or strongly supported. Do not include generic descriptions such as ``famous'', ``important'', or ``well-known''.\\[2pt]
Return a JSON list. Each item should contain:
\texttt{attribute}, \texttt{attribute\_type}, \texttt{evidence\_span}, and \texttt{confidence}.
\end{promptbox}
\caption{Prompt used to extract candidate attributes from a single webpage summary.}
\label{fig:web-attr-extract-prompt}
\end{figure}

\begin{figure}[H]
\centering
\begin{promptbox}[Cross-Page Attribute Aggregation Prompt]{promptgreen}
You are verifying candidate attributes extracted for the same entity from multiple webpages.\\
\textbf{Entity:} \{entity\_name\}\\
\textbf{Candidate attributes with sources:} \{attribute\_source\_list\}\\[2pt]
Group semantically equivalent attributes, count the number of distinct webpages supporting each group, and keep only attributes supported by at least three independent webpages. Remove attributes that are vague, promotional, contradictory, or not specifically about the target entity. For each retained attribute, output a concise canonical statement, its attribute type, supporting source ids, and a short reason explaining why it is stable enough for question construction.\\[2pt]
Return a JSON list with fields:
\texttt{canonical\_attribute}, \texttt{attribute\_type}, \texttt{supporting\_sources}, and \texttt{verification\_reason}.
\end{promptbox}
\caption{Prompt used to aggregate and verify repeated attributes across webpages.}
\label{fig:web-attr-aggregate-prompt}
\end{figure}

\clearpage

\section{Illustration of Exploration Degree}
\label{app:exploration-degree-example}

This section illustrates the exploration degree $\mathcal{E}(q)$ introduced in Section~\ref{sec:data-filtering} with a concrete example. The question is a text-to-image BrowseComp problem: textual browsing narrows the target entity and visual search supplies the final answer.

\paragraph{Question.}
\textit{``The 2021 International Eco-Design Award shortlisted several urban furniture projects. One shortlisted project was designed by the studio that later created the wayfinding system for the Northbank Wetland Center. Find the official logo of that studio and answer: what animal silhouette appears in the logo?''}
The ground-truth answer is \textbf{crane}.

\paragraph{Key evidence steps $\pi$.}
During construction, the pipeline derives a set of key evidence steps from the layered subgraph that underlies the question. These steps represent the core pieces of information that any successful trajectory should cover, but they are not a rigid chain: the subgraph structure may allow different browsing orders. For this question, the set $\pi$ is:
\begin{enumerate}[leftmargin=*,nosep]
    \item[$\rho_1$] Retrieve the shortlist of the 2021 International Eco-Design Award for urban furniture.
    \item[$\rho_2$] Among the shortlisted projects, identify the project designed by the studio that also created the Northbank Wetland Center wayfinding system.
    \item[$\rho_3$] Confirm the studio name as \textit{Marshline Design}.
    \item[$\rho_4$] Search for the official logo of \textit{Marshline Design}.
    \item[$\rho_5$] Inspect the logo and answer \textit{crane}.
\end{enumerate}
The set contains $|\pi|=5$ key steps. For a successful trajectory $\tau$, we measure its \emph{exploration length} $\ell(\tau,\pi)$ as the number of browsing actions (search / linksummary calls) it takes, normalized by $|\pi|$. Since $|\pi|$ is fixed per question, $\ell$ reflects how much browsing effort a successful route expends, regardless of the order in which the steps are covered.

\paragraph{Rollout examples.}
We present two successful rollouts obtained from the base agent. Both reach the correct answer and cover all five key evidence steps, but they differ substantially in how much browsing they require.

\begin{table}[H]
\centering
\small
\begin{tabularx}{\textwidth}{@{}c X c@{}}
\toprule
\textbf{Step} & \textbf{Trajectory A (direct route)} & \textbf{Matched $\rho$} \\
\midrule
1 & Search ``2021 International Eco-Design Award urban furniture shortlist.''          & $\rho_1$ \\
2 & Extract the list of shortlisted projects and studios from the award page.        & $\rho_1$ \\
3 & Search ``Northbank Wetland Center wayfinding system design studio.''              & $\rho_2$ \\
4 & Identify that Marshline Design created the wayfinding system.                    & $\rho_2,\rho_3$ \\
5 & Search ``Marshline Design official logo.''                                       & $\rho_4$ \\
6 & Inspect the logo image; the silhouette is a crane.                               & $\rho_5$ \\
\bottomrule
\end{tabularx}
\caption{Trajectory A: a compact browsing path with no dead ends. It takes $6$ browsing steps and covers all five key evidence steps, so $\ell = 6/5 = 1.2$.}
\label{tab:traj-A}
\end{table}

\begin{table}[H]
\centering
\small
\begin{tabularx}{\textwidth}{@{}c X c@{}}
\toprule
\textbf{Step} & \textbf{Trajectory B (reformulation-heavy route)} & \textbf{Matched $\rho$} \\
\midrule
1  & Search ``2021 International Eco-Design Award urban furniture winner.''            & $\rho_1$ \\
2  & Summarize the winner page; it does not list the shortlisted projects.            & $\rho_1$ \\
3  & Reformulate: search ``\dots urban furniture shortlist''; extract the projects.   & $\rho_1$ \\
4  & Search ``Northbank Wetland Center wayfinding system designer.''                  & $\rho_2$ \\
5  & Summarize an ambiguous page mentioning several studios (dead end).               & $\rho_2$ \\
6  & Reformulate: search ``\dots signage studio Marshline.''                          & $\rho_2$ \\
7  & Confirm Marshline Design created the wayfinding system.                          & $\rho_3$ \\
8  & Search ``Marshline Design official logo.''                                       & $\rho_4$ \\
9  & Inspect the logo image; the silhouette is a crane.                               & $\rho_5$ \\
10 & Emit the final answer: \textit{crane}.                                           & $\rho_5$ \\
\bottomrule
\end{tabularx}
\caption{Trajectory B: the agent reaches the same answer through a longer, reformulation-heavy route with a dead-end source before recovering. It takes $10$ browsing steps and also covers all five key evidence steps, so $\ell = 10/5 = 2.0$.}
\label{tab:traj-B}
\end{table}

Both trajectories reach the correct answer and cover the same key evidence steps, yet they differ substantially in exploration length ($\ell = 1.2$ for Trajectory~A versus $\ell = 2.0$ for Trajectory~B), reflecting genuinely different browsing behaviors.

\paragraph{Exploration degree of the example.}
Assume that for this question $q$ we sample $16$ rollouts and the subset of successful rollouts $\mathcal{Y}^{+}(q)$ contains trajectories with exploration lengths distributed similarly to the two examples above---some around $1.2$ (direct routes) and others around $2.0$ (reformulation-heavy routes). The exploration degree is the variance of $\ell$ across $\mathcal{Y}^{+}(q)$:
\begin{equation}
\mathcal{E}(q) = \mathrm{Var}_{\tau \in \mathcal{Y}^{+}(q)}\big[\ell(\tau,\pi)\big].
\end{equation}
A high $\mathcal{E}(q)$ indicates that successful trajectories vary substantially in browsing effort: some rollouts follow a compact direct route, while others reach the answer through longer reformulation-heavy or backtracking exploration. As discussed in Section~\ref{sec:data-filtering}, such instances are more likely to produce mixed-outcome groups (both successes and informative failures) during RL, providing rich contrastive signals for policy learning. In contrast, a problem where all correct trajectories require essentially the same browsing effort would yield near-zero variance, offering little behavioral contrast.